\documentclass{article}
\usepackage{spconf,amsmath,graphicx}
\usepackage{url}
\usepackage{amssymb}
\usepackage{multirow}

\usepackage{enumitem}
\setlist{nosep, leftmargin=14pt}

\usepackage{mwe} 

\title{MODALSURV: INVESTIGATING OPPORTUNITIES AND LIMITATIONS OF MULTIMODAL DEEP SURVIVAL LEARNING IN PROSTATE AND BLADDER CANCER}
\name{Noorul Wahab, Ethar Alzaid, Jiaqi Lv, Fayyaz Minhas, Adam Shephard, Shan E Ahmed Raza}
\address{TIA Centre, Department of Computer Science, University of Warwick, UK}
\begin{document}
%
\maketitle
\begin{abstract}
Accurate survival prediction is essential for personalised cancer treatment. We propose ModalSurv, a multimodal deep survival framework integrating clinical, MRI, histopathology, and RNA-sequencing data via modality-specific projections and cross-attention fusion. On the CHIMERA Grand Challenge datasets, ModalSurv achieved a C-index of 0.7402 (1st) for prostate and 0.5740 (5th) for bladder cancer. Notably, clinical features alone outperformed multimodal models on external tests, highlighting challenges of limited multimodal alignment and potential overfitting. Local validation showed multimodal gains but limited generalisation. ModalSurv provides a systematic evaluation of multimodal survival modelling, underscoring both its promise and current limitations for scalable, generalisable cancer prognosis. 
\end{abstract}
\section{Introduction}
\label{sec:intro}

Accurate prediction of patient survival and recurrence is central to computational oncology, guiding prognosis and treatment. Deep learning enables extraction of prognostic cues from histopathology, radiology, and transcriptomics, yet integrating these heterogeneous modalities remains difficult. Multimodal survival learning aims to capture complementary biological and clinical signals but is limited by scarce aligned datasets and the complexity of deep fusion models. 

Traditional WSI-based survival methods use CNN or multiple instance learning (MIL) frameworks to aggregate patch-level features \cite{YAO2020101789}, but they capture limited morphological context and overlook tumour microenvironment diversity. Foundation models pretrained on millions of histopathology tiles \cite{lu2021data, xu2024whole} now enable context-aware representations, though most applications remain unimodal, focusing on classification or risk estimation. 

Approaches such as CLAM  \cite{lu2021data} have improved WSI modelling through attention-based MIL, while Pathomic Fusion \cite{9186053} demonstrated benefits of combining histology with molecular or clinical data. However, most multimodal frameworks rely on feature concatenation or gated fusion, limiting cross-modal dependency learning, and few address time-to-event survival tasks.

We propose ModalSurv, a deep multimodal survival framework integrating foundation-model-derived pathology embeddings with clinical, transcriptomic, and MRI data via cross-attention fusion. This design enforces inter-modality learning by restricting self-attention within modalities. Evaluated on prostate and bladder cancer cohorts from the CHIMERA Grand Challenge \cite{chimeara}, ModalSurv achieved competitive C-indices and top leaderboard rankings. Results highlight both the potential and limitations of multimodal survival learning under data scarcity, where clinical features alone outperformed multimodal models, underscoring the need for larger, harmonised datasets and careful feature selection.

\section{MATERIALS AND METHODS}
\label{sec:methods}

\subsection{Datasets and Tasks}
\label{ssec:data}

We evaluated ModalSurv on two CHIMERA Grand Challenge tasks: biochemical recurrence prediction in prostate cancer (Task 1: clinical, WSI, MRI) and recurrence prediction in bladder cancer (Task 3: clinical, WSI, RNA-seq). Discovery cohorts included 95 patients for Task 1 (27 recurred, 68 censored) and 176 for Task 3 (64 recurred, 112 censored), with validation sets of 23 and 17 patients, respectively; test sets were held out. Survival times were measured in months. These small, partially aligned multimodal cohorts pose a challenging setting for deep fusion. Pretrained encoders were used only for feature extraction and were not fine-tuned. 

\subsubsection{Feature Extraction}
\label{sssec:featureextraction}

We incorporated four modalities to comprehensively represent each patient:

\begin{enumerate}
    \item Clinical variables:
Clinical variables were extracted from the provided JSON files. Mixed-format categorical fields (e.g., ‘12a’) were numerically encoded, all values were cast to float, and missing entries were removed to produce clean tabular inputs. This yielded 10 features for Task 1 and 8 for Task 3. Several variables (such as stage, grade, lymphovascular invasion, or BCG-response subtype) already embed information derived from other assessments, meaning that even ‘clinical’ data may implicitly contain multimodal context.
    \item Whole-Slide Image (WSI) features:
Histopathology slides were processed using the Trident \cite{zhang2025accelerating} pipeline with CONCH \cite{lu2024visual} tile-level embeddings (10× magnification, 1024-pixel tiles). The Titan foundation model \cite{ding2025multimodal} was used for WSI-level representations (768-dimensions), providing context-rich embeddings that capture global tissue architecture beyond traditional CNN features. Tissue segmentation was performed with GrandQC to exclude background regions. For patients with multiple WSIs, the alphabetically first slide was used for consistency. Because of resource constraints on the challenge server, a maximum of 500 patches was randomly selected from each slide.
    \item MRI features:
Multiparametric MRI volumes (e.g., T2-weighted, DWI, ADC) were processed using a pretrained MedicalNet ResNet-50 \cite{chen2019med3d}, to obtain volumetric embeddings. Images were resampled to the mask space, cropped or padded to 19 axial slices, resized to 128×120 pixels, and z-normalised. Tumour masks, when available, were applied to remove non-tumoral voxels. The resulting 19×128×120 tensors were passed through MedicalNet, and global average pooling produced 2048-dimensional embeddings.
    \item RNA-seq features:
RNA sequencing data comprising 19,359 genes were reduced to 128 principal components using principal component analysis (PCA) to mitigate redundancy and reduce dimensionality.
\end{enumerate}

\subsubsection{Preprocessing and Data Splits}
\label{sssec:preprocessing}
Each modality was independently z-score normalised using training-fold statistics, with the same scalers applied to validation and test sets to prevent leakage. We used 5-fold stratified cross-validation to maintain recurrence–censoring ratios, and repeated all experiments ten times with different seeds to capture optimisation variability. Fold assignments were fixed and shared across modalities to ensure reproducibility and consistent ensembling.

\subsection{Model Architecture}
\label{ssec:model_archi}

\begin{figure}[htb]

  \centering
  \centerline{\includegraphics[width=8.5cm]{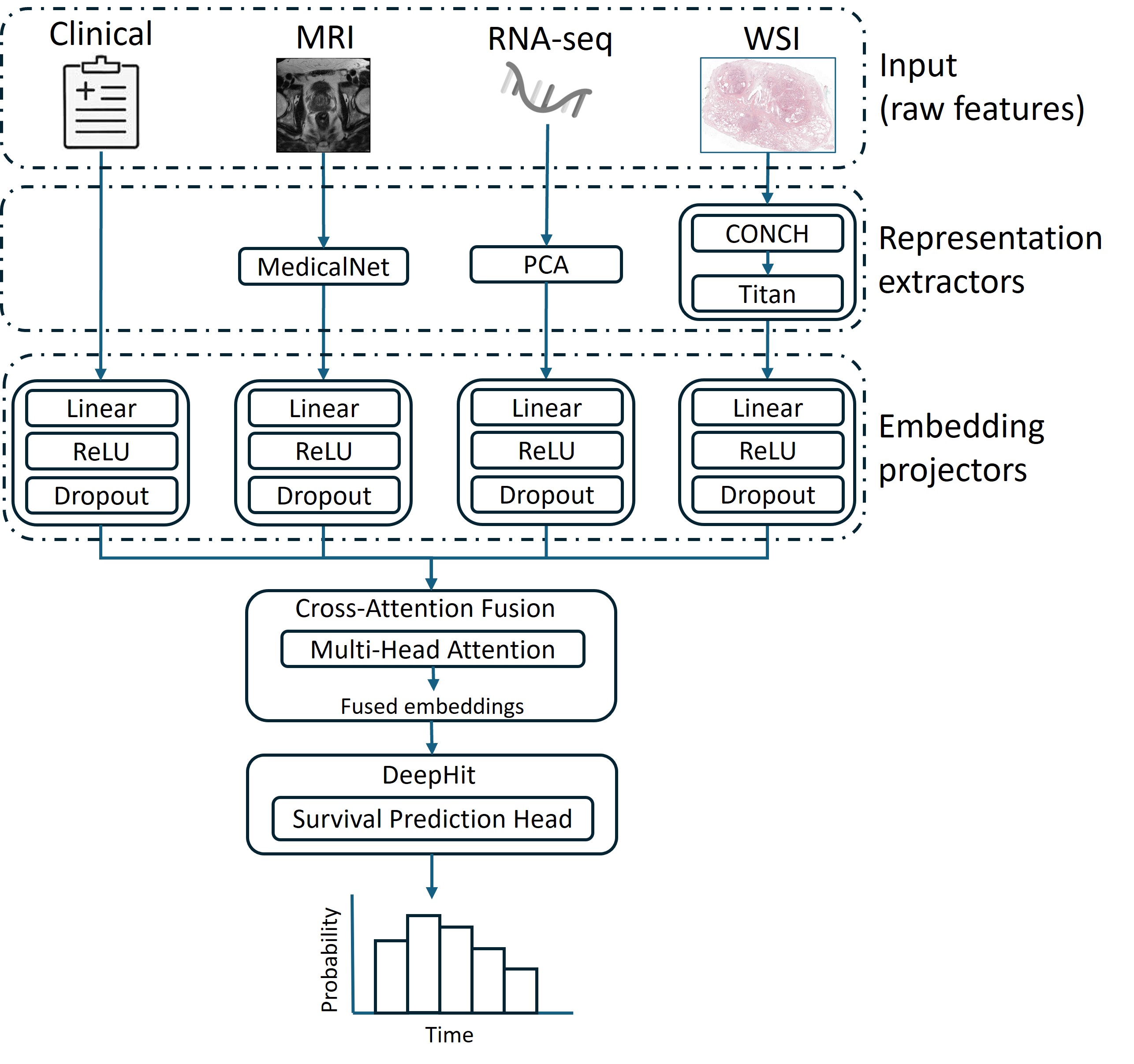}}
  \caption{ModalSurv: Overview of the ModalSurv pipeline integrating multimodal clinical, MRI, RNA-seq, and WSI features through modality-specific encoders, cross-attention fusion, and a DeepHit-based survival prediction head.}
\label{fig:pipeline}
\end{figure}

ModalSurv extends the DeepHit \cite{Lee_Zame_Yoon_van} survival framework to handle multiple heterogeneous input modalities. The architecture consists of three main components (Fig. ~\ref{fig:pipeline}):

\begin{enumerate}
    \item Modality-specific encoders:
Each available modality (clinical, MRI, WSI, RNA) is passed through a small projection head comprising a linear layer followed by a ReLU activation to  produce a compact 128-dimensional latent embedding. These projections map diverse input features into a common latent space suitable for joint learning.

    \item Fusion mechanism:
Each modality embedding $f_i \in \mathbb{R}^D$ is treated as a query that attends to the other modalities (no self-attention). For each modality we compute an attended vector via a cross-attention block:
$$
\tilde{f}_i \;=\; \mathrm{CrossAttn}\big(f_i,\{f_j\}_{j\neq i}\big),
$$
where each CrossAttn block applies multi-head attention followed by a residual feed-forward network and layer normalisation. The attended vectors are stacked and passed to a gating network that produces per-modality soft weights $w_i$ for each sample:
$$
w \;=\; \mathrm{Softmax}\big(\mathrm{Gate}([\tilde{f}_1,\ldots,\tilde{f}_N])\big), \qquad w_i \ge 0,\ \sum_{i} w_i = 1.
$$
The fused representation is the weighted sum of attended vectors, followed by a final linear projection:
$$
f_{\mathrm{fused}} \;=\; \mathrm{Proj}\!\left(\sum_{i=1}^{N} w_i \,\tilde{f}_i\right).
$$
Dropout is applied after projection for regularisation.

    \item Survival prediction head and objective:
The fused embedding is fed into a shallow DeepHit head, a linear layer with SoftMax activation that produces a discrete-time probability mass function (PMF) over survival intervals. Although the prediction head is simple, the overall model remains deep due to the nonlinear projection and cross-attention fusion. The network is trained with the DeepHit loss \cite{Lee_Zame_Yoon_van}, which combines a likelihood term for event-time estimation and a ranking term for risk ordering:

$\mathcal{L} = (1 - \alpha)\mathcal{L}_{\text{likelihood}} + \alpha \mathcal{L}_{\text{ranking}}$

where $\alpha \in [0,1]$ balances discrimination and calibration. The likelihood handles censoring, while the ranking term penalizes incorrect risk ordering.

\end{enumerate}

Due to limited training data, submission constraints, and risk of overfitting, we submitted the clinical-only ModalSurv model for Task 1. For Task 3, despite clinical features performing best on the discovery set, a multimodal (clinical + WSI) model was submitted to assess test-set generalisation. WSI inputs were restricted to 500 patches per patient by challenge server limits.

\subsection{Training Procedure}
\label{ssec:training}

For both tasks, models were trained independently for each cross-validation fold using the Adam optimiser with an initial learning rate of $1 \times 10^{-3}$. Early stopping based on validation C-index was applied to mitigate overfitting. Regularisation included dropout (p=0.3) and L2 penalties in projection layers. Survival times were discretised into 30 equal-width bins covering the observed range of follow-up durations for each task. The WSI feature extraction was performed on an 80 GB GPU of a DGX A100, whereas model training was done on a Precision 5820 Tower CPU.

Given the small dataset sizes, all hyperparameters were shared across modalities and tasks to reduce model complexity and prevent overfitting.

\subsection{Ensembling, Inference, and Evaluation}
\label{ssec:ensemble}

During inference, predictions from fold-specific models were averaged to form an ensemble survival distribution. Expected survival times were computed as the weighted sum of interval midpoints and probabilities. Model performance was evaluated using the concordance index (C-index), measuring agreement between predicted and observed event times. For CHIMERA Challenge submission, the trained ModalSurv model was containerised with Docker and run on blind test sets via the official platform.


\section{RESULTS}
\label{sec:results}
Tables 1 and 2 present the C-indices for the baseline Cox Proportional Hazards (Cox PH) model and the proposed ModalSurv across 5-fold cross-validation (discovery), CHIMERA validation, and test sets for Task 1 (prostate cancer recurrence) and Task 3 (bladder cancer recurrence), respectively.

\subsection{Task 1: Prostate Cancer (Biochemical Recurrence Prediction)}
\label{ssec:task1}

Clinical features alone provided a strong prognostic baseline (C-index 0.829 ± 0.06 for CoxPH; 0.843 ± 0.08 for ModalSurv). MRI features performed worse (0.682 ± 0.07), while WSI embeddings achieved similar performance (0.846 ± 0.09), confirming prognostic value in tissue morphology. The clinical + WSI model with CrossAtt achieved the best discovery performance (0.891 ± 0.01) but dropped notably on validation (0.645), suggesting overfitting from the small cohort, patch sampling, and domain shifts. The same feature combination via simple contatenation resulted in a C-index of 0.850 ± 0.02. On the CHIMERA leaderboard, top models relied on clinical features alone, underscoring their robustness over complex multimodal fusion.

\begin{table}[t]
\centering
\caption{\textbf{Performance of ModalSurv for time-to-biochemical recurrence prediction (Task 1).}
C-index (mean ± standard deviation) for 5-fold discovery and challenge validation (Val.) and test sets.
Clin (Clinical). Methods for 2$^{nd}$ and 3$^{rd}$ Pos. not published yet.}
\resizebox{\columnwidth}{!}{%
\begin{tabular}{|l|l|c|c|c|}
\hline
\textbf{Model} & \textbf{Features} & \textbf{Discovery} & \textbf{Val.} & \textbf{Test} \\ \hline

\multirow{1}{*}{Cox PH}
& Clin & 0.829$\pm$0.06 & - & - \\ \hline

\multirow{8}{*}{ModalSurv}
& Clin & 0.843$\pm$0.08 & \textbf{0.8182} & \textbf{0.7402} \\ 
& MRI & 0.682$\pm$0.07 & - & - \\
& WSI & 0.846$\pm$0.09 & - & - \\
& Clin+MRI & 0.791$\pm$0.10 & - & - \\
& MRI+WSI & 0.804$\pm$0.12 & - & - \\
& Clin+WSI & \textbf{0.891$\pm$0.01} & 0.6446 & - \\
& Clin+MRI+WSI & 0.833$\pm$0.11 & - & - \\ \hline

2$^{nd}$ Pos. & Clin & - & 0.7521 & 0.7294 \\ \hline
3$^{rd}$ Pos. & Clin & - & 0.7355 & 0.7280 \\ \hline

\end{tabular}
}
\label{tab:task1_table}
\end{table}

\subsection{Task 3: Bladder Cancer (Recurrence Prediction)}
\label{ssec:task3}

For bladder cancer, clinical features again provided the strongest signal (C-index 0.684 ± 0.04 for CoxPH; 0.749 ± 0.08 for ModalSurv). WSI-only (0.649 ± 0.06) and RNA-seq-only (0.664 ± 0.08) models performed worse, reflecting challenges in modelling high-dimensional data with limited samples. The clinical + WSI model improved cross-validation (0.733 ± 0.06) but dropped on external validation (0.4565) and test (0.5740), indicating overfitting and poor generalisation. Leaderboard results confirmed this trend, with clinical-only models achieving more stable performance (up to 0.6828) compared to multimodal approaches showing high variance (0.8478–0.5977).

\begin{table}[t]
\centering
\caption{\textbf{Performance of ModalSurv for time-to bladder cancer recurrence prediction (Task 3).} 
C-index (mean ± standard deviation) for 5-folds discovery and challenge validation (Val.) and test sets. 
Clin (Clinical). Methods for 1$^{st}$ to 4$^{th}$ Pos. not published yet.}
\resizebox{\columnwidth}{!}{%
\begin{tabular}{|l|l|c|c|c|}
\hline
\textbf{Model} & \textbf{Features} & \textbf{Discovery} & \textbf{Val.} & \textbf{Test} \\ \hline

Cox PH & Clin & 0.684$\pm$0.04 & 0.5870 & - \\ \hline

1$^{st}$ Pos. & Clin & - & 0.500 & \textbf{0.6828} \\ \hline
2$^{nd}$ Pos. & Clin & - & 0.6169 & 0.6792 \\ \hline
3$^{rd}$ Pos. & Clin & - & 0.7826 & 0.6082 \\ \hline
4$^{th}$ Pos. & Clin & - & \textbf{0.8478} & 0.5977 \\ \hline

\multirow{8}{*}{ModalSurv}
& Clin & \textbf{0.749$\pm$0.08} & - & - \\
& RNA & 0.664$\pm$0.08 & - & - \\
& WSI & 0.649$\pm$0.06 & - & - \\
& Clin+RNA & 0.689$\pm$0.07 & - & - \\
& RNA+WSI & 0.657$\pm$0.05 & - & - \\
& Clin+WSI & 0.733$\pm$0.06 & 0.4565 & 0.5740 \\
& Clin+MRI+WSI & 0.690$\pm$0.05 & - & - \\ \hline
\end{tabular}
}
\label{tab:task3_table}
\end{table}

\section{DISCUSSION AND CONCLUSION}
\label{sec:discussion}
This study systematically evaluated multimodal deep survival modelling for prostate and bladder cancer recurrence using the CHIMERA challenge datasets. ModalSurv integrates clinical, radiological, histopathological, and molecular data through cross-attention fusion to capture complementary prognostic signals. Across both tasks, clinical features provided the most stable and generalisable performance, likely because they already encode multimodal information such as grade, stage, and biomarker-based subtypes. Although multimodal fusion improved cross-validation results, its performance declined on unseen data, suggesting overfitting under limited-sample conditions.

These findings underscore a key challenge in multimodal survival learning: while deep fusion can leverage complementary cues, its generalisation remains limited by small, imbalanced, and misaligned datasets. In addition, the limited validation data available for model selection may have biased performance estimates. Foundation-model-derived WSI features capture rich morphology, but integrating them effectively with molecular and imaging data requires larger, harmonised cohorts and consistent preprocessing.

Overall, ModalSurv demonstrates both the promise and current barriers of multimodal survival modelling. Clinical-only models remain strong baselines when data are scarce, while cross-attention fusion offers a scalable direction once larger and better-curated datasets are available. Future work should emphasise uncertainty-aware fusion, adaptive modality weighting, and multi-institutional validation to ensure clinically robust multimodal survival prediction.

\section{Acknowledgments}
\label{sec:acknowledgments}

Thanks to Salinder Tandi for their system administrative help, to TIA Centre and Department of Computer Science, University of Warwick for providing computational resources and to CHIMERA Challenge organisers for organising the challenge.

\bibliographystyle{IEEEbib}
\bibliography{references}

\end{document}